\documentclass{article}

\usepackage[utf8]{inputenc} 
\usepackage[T1]{fontenc}   
\usepackage{times}          

\usepackage{amsmath}
\usepackage{amsfonts}
\usepackage{nicefrac}    

\usepackage[pdftex]{graphicx}
\graphicspath{{./pdf/}}
\usepackage{adjustbox}
\usepackage{placeins}

\usepackage{booktabs}   
\usepackage{multirow}
\usepackage{tabularx}

\usepackage{enumitem}
\usepackage{listings}
\usepackage{microtype}

\usepackage[dvipsnames]{xcolor}

\usepackage[numbers]{natbib}
\usepackage{doi}
\usepackage{xurl}        
\usepackage{url}
\usepackage{hyperref}     

\usepackage{csagh}       
\makeatletter
\def\enditemize{%
  \endlist
  \global\@itemdepth\z@ 
}
\def\endenumerate{%
  \endlist
  \global\@enumdepth\z@ 
}
\makeatother

\usepackage{lipsum}        

\graphicspath{{./pdf/}}

\begin{document}
\begin{opening}

\title{Bielik 7B v0.1: A Polish Language Model – Development, Insights, and Evaluation}

\author[SpeakLeash, Azurro, krzysztof.ociepa@bielik.ai]{Krzysztof Ociepa}
\author[SpeakLeash, ACK Cyfronet AGH, lukasz.flis@cyfronet.pl]{\L{}ukasz Flis}
\author[SpeakLeash, Jagiellonian University, Enelpol, krzysztof.pawel.wrobel@uj.edu.pl]{Krzysztof Wr\'obel}
\author[SpeakLeash, ACK Cyfronet AGH, adrian.gwozdziej@cyfronet.pl]{Adrian Gwo\'zdziej}
\author[SpeakLeash, remigiusz.kinas@bielik.ai]{Remigiusz Kinas}

\received{TODO}
\revised{TODO}
\accepted{TODO}

\begin{abstract}
We introduce Bielik 7B v0.1, a 7-billion-parameter generative text model for Polish language processing. Trained on curated Polish corpora, this model addresses key challenges in language model development through innovative techniques. These include Weighted Instruction Cross-Entropy Loss, which balances the learning of different instruction types, and Adaptive Learning Rate, which dynamically adjusts the learning rate based on training progress. To evaluate performance, we created the Open PL LLM Leaderboard and Polish MT-Bench, novel frameworks assessing various NLP tasks and conversational abilities. Bielik 7B v0.1 demonstrates significant improvements, achieving a 9 percentage point increase in average score compared to Mistral-7B-v0.1 on the RAG Reader task. It also excels in the Polish MT-Bench, particularly in Reasoning (6.15/10) and Role-playing (7.83/10) categories. This model represents a substantial advancement in Polish language AI, offering a powerful tool for diverse linguistic applications and setting new benchmarks in the field.
\begin{center}
\small\textit{The paper has been accepted for publication in Computer Science journal:
\url{http://journals.agh.edu.pl/csci}.}
\end{center}
\end{abstract}


\end{opening}

\section{Introduction}
The rapid advancement in natural language processing (NLP) has led to the development of increasingly sophisticated language models that can understand and generate human-like text. These models have shown remarkable success in various linguistic tasks across multiple languages. However, the development of high-performing models for less-resourced languages remains a significant challenge due to the scarcity of large and diverse datasets and computational resources.

Existing Polish language models, such as TRURL 2 \cite{trurl} and Qra \cite{qra}, have made important strides in this domain. TRURL 2, a collection of fine-tuned Llama 2 models with 7 billion and 13 billion parameters was trained on approximately 1 million conversational Polish and English samples, with a context size of 4,096 tokens. Another series of models is Qra, which comprises continuously pretrained models with 1, 7, and 13 billion parameters. The Qra models were trained on Polish data, totaling 90 billion tokens, and also employ a context size of 4,096 tokens. While numerous other Polish-focused language models exist, the majority of them are fine-tuned using significantly smaller datasets or fine-tuning approaches, which can limit their performance and versatility.

This paper introduces Bielik 7B v0.1, a state-of-the-art Polish language model developed as a collaborative effort between the open-science project SpeakLeash and the High Performance Computing (HPC) center: ACK Cyfronet AGH. Bielik 7B v0.1 is an evolution of the Mistral 7B v0.1 model \cite{jiang2023mistral7b}, enhanced to understand and generate Polish text with high accuracy. This model leverages a massive corpus of Polish texts and advanced machine learning techniques, making it a pioneering tool in the realm of Polish Natural Language Processing (NLP). The development of Bielik 7B v0.1 addresses several challenges, including the adaptation of a model trained primarily on English data to the Polish language, which involves significant linguistic and semantic adjustments.

In the following sections, we detail the architecture of Bielik 7B v0.1, describe the dataset preparation, discuss the training process, and evaluate the model's performance on various NLP tasks. Our results demonstrate that Bielik 7B v0.1 not only advances the state of Polish language understanding but also serves as a valuable resource for further research and application in Polish NLP.

\section{Model and Tokenizer}

In this section, we introduce the model design and tokenizer, presenting architectural decisions and configurations.

\subsection{Model Architecture}

\begin{table}[h]
\centering
\begin{tabular}{ll}
\toprule
Parameter & Value \\
\midrule
Layers & 32 \\
Model Dimension & 4096 \\
Attention Heads & 32 \\
Key/Value Heads & 8 \\
Head Size & 128 \\
Intermediate Size & 14336 \\
Activation Function & SwiGLU \\
Vocabulary Size & 32000 \\
Positional Embeddings & RoPE ($\theta = 10000$) \\
Context Length & 8192 \\
Sliding Window & 4096 \\
\bottomrule
\end{tabular}
\caption{Model architecture.}
\label{tab:model-architecture}
\end{table}

\noindent The Bielik 7B v0.1 model builds upon the Transformer architecture \cite{Vaswani2017AttentionIA}, with its key parameters detailed in Table \ref{tab:model-architecture}, and incorporates a suite of advanced techniques to enhance its performance.

\noindent \textbf{Self-attention with causal masks} \cite{Vaswani2017AttentionIA} allows the model to weigh the importance of different parts of the input sequence. The causal mask ensures that the model only attends to previous tokens, which is crucial for maintaining the autoregressive property in language modeling tasks.

\noindent \textbf{Grouped-query attention (GQA)} \cite{ainslie-etal-2023-gqa} reduces computational complexity and memory usage while maintaining model quality. It achieves this by using fewer key-value heads than query heads, allowing for more efficient processing of long sequences.

\noindent \textbf{Sliding Window Attention} \cite{Child2019GeneratingLS,Beltagy2020LongformerTL} limits the attention span to a fixed window size, reducing computational complexity from quadratic to linear in sequence length. It enables the model to process longer sequences more efficiently while still capturing local context effectively.

\noindent \textbf{SwiGLU activation function} \cite{Dauphin2016LanguageMW,shazeer2020gluvariantsimprovetransformer} is a combination of the Swish activation function and Gated Linear Units (GLU), offering improved performance and trainability compared to traditional activation functions like ReLU.

\noindent \textbf{Rotary Positional Embeddings (RoPE)} \cite{SU2024127063} allow the model to better capture the relative positions of tokens in the input sequence, offering advantages over absolute positional embeddings. It excels in tasks requiring nuanced understanding of token positions, providing better extrapolation to longer sequences and improving overall performance.

\noindent \textbf{Root Mean Square Layer Normalization (RMSNorm)} \cite{10.5555/3666122.3668105} is used for normalizing activations within the network. It offers improved training stability and slightly faster computation compared to traditional Layer Normalization, contributing to more efficient training and inference.

\noindent \textbf{Pre-normalization} applies layer normalization before the self-attention and feed-forward layers, rather than after, resulting in improved model convergence and overall performance.

The Bielik 7B v0.1 model was adapted from the Mistral 7B v0.1 model and further pretrained. The decision to use an existing model instead of training our own from scratch was due to the lack of access to sufficient high-quality data. Additionally, training from scratch would have required significantly more resources, including GPU power and time. We chose the Mistral 7B v0.1 model because of its strong performance in benchmarks and its permissive Apache 2.0 license.

\subsection{Tokenizer}
\begin{table*}[h]
\centering
\begin{tabular}{lll|lll|lll}
\toprule
                    &  Vocab               &  Avg              & \multicolumn{3}{c|}{Polish} & \multicolumn{3}{c}{English} \\
Tokenizer           &  size &  tokens & Tokens   & CpT    & TpW    & Tokens    & CpT    & TpW    \\
APT3                & 31980           & 480            & 344      & 5.22   & 1.48   & 615       & 3.15   & 1.93   \\
Llama2              & 32000           & 554            & 681      & 2.63   & 2.94   & 427       & 4.53   & 1.34   \\
Mistral v0.1        & 32000           & 578            & 747      & 2.40   & 3.22   & 408       & 4.75   & 1.28   \\
Llama2 + APT3       & 57362           & 442            & 441      & 4.07   & 1.90   & 442       & 4.38   & 1.39   \\
Mistral v0.1 + APT3 & 58690           & 450            & 493      & 3.64   & 2.12   & 407       & 4.76   & 1.28    \\
\bottomrule
\end{tabular}
\caption{Comparison of token count, characters per token (CpT), and tokens per word (TpW) for the preamble of the Constitution of the Republic of Poland in Polish and English versions, processed by various tokenizers: APT3 (dedicated Polish language tokenizer), Llama2 and Mistral v0.1 (multilingual tokenizers with minimal Polish support), and merged tokenizers Llama2 + APT3 and Mistral v0.1 + APT3.}
\label{tab:tokenizers-comparison}
\end{table*}

One measure of the effectiveness of the tokenization process is the count of tokens generated for the input text. A lower number of tokens indicates faster and more efficient text generation by the language model. The tokenizer from the Mistral 7B model was not specifically trained for the Polish language. Therefore, we conducted a series of experiments aimed at expanding the original tokenizer to include Polish tokens. Our approach to expanding the tokenizer involved incorporating tokens from the Polish APT3 model \cite{AzurroAPT3Base1B} by extending the model's edge layers (embeddings and language model head) and continuing the training process. We chose the preamble of the Constitution of the Republic of Poland as the benchmark text because it effectively captures the essence of Polish writing and includes official English versions for comparative analysis. Table \ref{tab:tokenizers-comparison} presents a detailed comparison of various metrics, including token count, characters per token (CpT), and tokens per word (TpW). These metrics illustrate the performance of different tokenizers when applied to both the Polish and English versions of the preamble.

Despite achieving good results on benchmarks with the trained models, we observed issues in text generation, which occasionally manifested as incorrect token combinations for Polish words. This problem arose partly due to the ambiguity that occurs when merging pairs of tokens during the tokenization process \cite{imamura2022extendingsubwordingmodelmultilingual}. This process utilizes the byte pair encoding (BPE) algorithm \cite{sennrich2016neuralmachinetranslationrare}, which is implemented through SentencePiece \cite{kudo2018sentencepiecesimplelanguageindependent}. Since the tokens from the APT3 model tokenizer and the Mistral 7B model tokenizer are not mutually exclusive (their vocabularies overlap), ambiguity arises during the merging of token pairs, making it impossible to directly combine both tokenizers.

In light of these issues, we decided to retain the original tokenizer from the Mistral 7B model, which has a vocabulary size of 32,000 tokens, while continuing to explore potential expansion options for future model versions.

\section{Pre-training}
The primary objective of the pre-training phase was to enhance the model's Polish language capabilities, focusing on both accuracy and fluency. To accomplish this, we employed a diverse selection of high-quality Polish texts. These materials were subjected to rigorous cleaning procedures and meticulous quality evaluations, ensuring the highest standard of training data.

\subsection{Pre-training Data} \label{Pre-training-Data}
The pre-training of the Bielik model involved constructing a novel, diverse, and high-quality dataset, primarily made up of Polish language texts. We leveraged resources from the SpeakLeash project \cite{speakleashorg}. Using metadata assigned to each document, which included information about its topic and various stylometric features, we selected 18 million documents from different datasets that offered high quality and topic diversity. These selected texts underwent thorough cleaning and quality assessment procedures, detailed in sections \ref{Data-Cleanup} and \ref{Quality-Evaluation}. Additionally, we removed documents where, although robots.txt did not prohibit scraping, the terms and conditions explicitly forbade using them for training language models. Only documents meeting our stringent quality criteria were retained for training and subsequently tokenized. This meticulous process yielded a training dataset comprising 22 billion tokens. To improve the model's adaptation to a new language and mitigate catastrophic forgetting \cite{Li2022OvercomingCF,pmlr-v199-ostapenko22a,ibrahim2024simplescalablestrategiescontinually}, we supplemented our training dataset with English texts, sourced from the SlimPajama dataset \cite{cerebras2023slimpajama}, known for its diverse and high-quality English content. Ultimately, our final training dataset consisted of 36 billion tokens.
\subsubsection{Data Cleanup} \label{Data-Cleanup}
The foundation of our pre-training corpus was a broad collection of texts from the Polish web (including processed data from the CulturaX and HPLT datasets), supplemented with digitized library resources and publicly available documents. To improve the quality of this data, we implemented a series of heuristics aimed at removing damaged and unwanted text fragments, anonymizing personal data (such as physical addresses, email addresses, phone numbers, and URLs), and fixing encoding or formatting issues. As a result of this process, we obtained higher-quality texts, which were ready for a detailed quality assessment.

\subsubsection{Quality Evaluation} \label{Quality-Evaluation}
\begin{figure}[ht]
\centering
\includegraphics[width=\columnwidth]{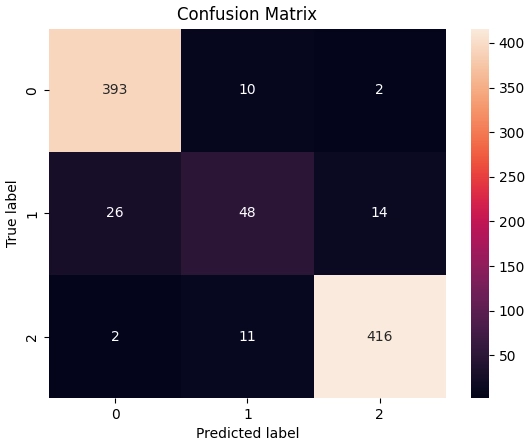}
\caption{Confusion matrix illustrating validation results for the XGBoost classifier model.}
\label{fig:XGBoost-validation}
\end{figure}

To create the training dataset for text quality evaluation, we manually annotated 9,000 documents, assigning each to one of three quality classes: HIGH, MEDIUM, or LOW. To ensure high internal consistency in the evaluation, the entire labeling process was carried out by a single, specialized annotator. The classification criteria were defined as follows:
\begin{itemize}
    \item \textbf{HIGH class} included documents of high substantive value, characterized by clear, logical, and well-formatted text. Minor formatting errors were permissible if they did not affect readability (e.g., in valuable content from internet forums or official documents).
    \item \textbf{LOW class} comprised clearly problematic texts containing encoding errors, broken formatting (e.g., incorrectly converted tables), thematically mixed-up or truncated fragments, as well as vulgar content or texts consisting mainly of non-linguistic data (e.g., financial reports).
    \item \textbf{MEDIUM class} served as a buffer for documents of a mixed nature, which contained both valuable fragments and significant flaws (e.g., an article snippet surrounded by website interface elements like menus or headers). This class identified texts with potential for future recovery, for instance, through automated correction.
\end{itemize}
The lower prevalence of the MEDIUM class in the manually annotated dataset (as shown in Figure \ref{fig:XGBoost-validation}) stemmed from both its natural rarity in the source data and the fact that the annotation process prioritized creating a balanced set for the clearly defined HIGH and LOW classes, whose assessment was less time-consuming.

For each document, we calculated 266 stylometric features, including metrics such as the frequency of verbs, nouns, sentences, and punctuation marks. This comprehensive set of features was derived based on the methodology outlined in the StyloMetrix tool \cite{okulska2023stylometrixopensourcemultilingualtool}. These linguistic and structural attributes provided a multifaceted representation of each text's stylistic properties.

Using these stylometric features as input, we trained an XGBoost classifier model. This machine learning approach allowed us to leverage the complex interactions between various textual characteristics to predict document quality effectively, as presented in Table \ref{tab:XGBoost-validation} and Figure \ref{fig:XGBoost-validation}.
\begin{table}[htbp]
\centering
\begin{tabular}{ccc}
\toprule
\textbf{Precision} & \textbf{Recall} & \textbf{F1} \\
\midrule
0.8640 & 0.8285 & 0.8431 \\
\bottomrule
\end{tabular}
\caption{Validation results for the XGBoost classifier model.}
\label{tab:XGBoost-validation}
\end{table}

To determine an appropriate threshold for identifying high-quality documents, we conducted a manual analysis of 1,000 documents. Based on this thorough examination, we established a cut-off point for the HIGH category at a probability score exceeding 90\%. Documents that did not meet this threshold were excluded from the target training dataset of the Bielik model.

\subsection{Training Hyperparameters}
\begin{figure}[h]
\centering
\includegraphics{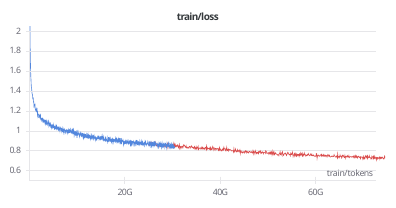}
\caption{Training loss over the training tokens for the base model.}
\label{fig:base-training-loss}
\end{figure}

\begin{figure}[h]
\centering
\includegraphics{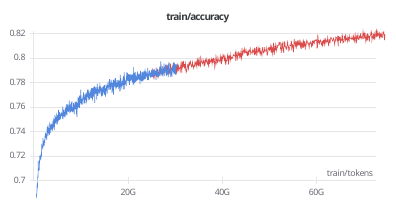}
\caption{Training accuracy over the training tokens for the base model.}
\label{fig:base-training-acc}
\end{figure}

We utilized the AdamW optimizer \cite{Loshchilov2017DecoupledWD} with hyperparameters $\beta_1 = 0.9$, $\beta_2 = 0.95$, and weight decay = 0.1. The learning rate followed a cosine decay schedule, starting at 3e-05 and decreasing to 2e-05, with a warmup period of 2000 iterations. Training continued for a total of 17350 iterations. We employed a global batch size of 256, composed of local batches of size 4. The gradient clipping norm was set to 1.0, and we used mixed precision with bfloat16. The model was trained on 36B tokens over 2 epochs, with a maximum context length of 4096. The training loss and accuracy over the training tokens for the base model are presented in Figures \ref{fig:base-training-loss} and \ref{fig:base-training-acc}.

\section{Post-training}
After finishing the pre-training phase, we moved on to the post-training phase, which focused on improving the model's capabilities across various areas, such as coding, mathematics, logical reasoning, and following instructions.

\subsection{Post-training Data} \label{Post-training-Data}
A significant challenge in developing high-performing models for the Polish language is the scarcity of large-scale, open-source instruction datasets. To address this gap, we initiated the creation of a comprehensive Polish instruction-following dataset, which is continuously expanded and refined. Our approach combines manually authored data with a large-scale automated generation pipeline, supplemented by high-quality English datasets.

\subsubsection{Polish Instruction Dataset Creation}
Our Polish dataset consists of two primary components:

\textbf{Manually Authored Instructions.} A high-quality set of instructions was developed by our annotators. This collection focuses on tasks that require deep linguistic understanding or can be programmatically verified. The categories include, among others: text classification, sentiment analysis, and natural language processing tasks such as part-of-speech tagging (e.g., identifying verbs or nouns using libraries like Morfeusz and spaCy).

\textbf{Automatically Generated Instructions.} To augment the manually curated data, we developed a large-scale automatic generation pipeline. We selected a diverse collection of 1 million high-quality articles from our pre-training corpus (see Section \ref{Pre-training-Data}). Using the Mixtal 8x22B model, we generated instruction-response pairs for each article, covering tasks such as summarization, question answering based on the provided context, and email composition.

\subsubsection{Quality Assurance and Data Composition}
To ensure the quality of the automatically generated data, we employed an LLM-as-a-judge methodology for large-scale evaluation. To further validate this automated assessment, approximately 1,000 instruction-response pairs were spot-checked manually. These carefully verified examples now serve as a "gold standard" set, which can be used for future evaluation and to guide further data generation.

To further increase the number and diversity of instructions, we utilized publicly accessible collections of English instructions, such as the OpenHermes-2.5 \cite{OpenHermes2.5} and orca-math-word-problems-200k \cite{mitra2024orcamath} datasets. These English-language resources accounted for approximately half of the instructions used in the final training mixture.

As a result, we compiled a final training dataset containing over 2.3 million instructions, amounting to more than 700 million tokens. To support transparency and enable further research in the Polish language domain, we plan to release a representative portion of our Polish instruction dataset in the future.

\subsection{Supervised Fine-Tuning}
The varying quality of training instructions negatively impacts a model's benchmark performance, as demonstrated in previous studies, which found that poor-quality instructions degrade model capabilities \cite{NEURIPS2023_ac662d74}. These studies showed that smaller, higher-quality instruction datasets often yield better results than larger, noisier datasets. To address this, we introduced several improvements, summarized below, while still utilizing the previously mentioned datasets.

\subsubsection{Masked Tokens}
We employed a masked token approach, selectively applying loss only to certain parts of the output. Specifically, we masked the loss on user instruction and control tokens \cite{shi2024instructiontuninglossinstructions}. This technique ensures that these tokens do not contribute to the overall loss during training, allowing the model to focus on learning from the content tokens.

\subsubsection{Adaptive Learning Rate} \label{ALR}
The lengths of instructions can vary significantly, leading to fluctuations in the number of tokens used in computing the loss function. To ensure consistent influence from each instruction, we implemented an adaptive learning rate (ALR). This approach is based on prior research that links learning rates to batch sizes \cite{Granziol2020LearningRA}. In particular, the learning rate (LR) is scaled according to the square root of the ratio between the number of tokens in the batch (T) and the baseline batch size (BS):
\begin{equation}
\text{ALR} = \text{LR} \cdot \sqrt{\frac{\text{T}}{\text{BS}}}
\end{equation}

\subsubsection{Weighted Instruction Cross-Entropy Loss} \label{WICEL}
This strategy, inspired by weighted cross-entropy loss \cite{Wu2024MisclassificationguidedLU}, offline reinforcement learning \cite{pmlr-v162-xu22l} and C-RLFT \cite{wang2024openchatadvancingopensourcelanguage}, enabled us to effectively utilize mixed-quality training data annotated with fine-grained weight labels. 


Given the SFT conversation dataset $\mathcal{D} = {(x_i, y_i)}$, where $x_i$ indicates the instruction, $y_i$ is its corresponding response, we assign a weight $w_i \in (0, 1]$ to each instruction-response pair $(x_i, y_i)$ that is representing the quality of the pair. This allows us to construct a weighted dataset, $\mathcal{D}_w$, where the highest quality pairs are assigned a weight of $1.0$, while lower quality instructions have smaller weights ($w_i < 1.0$). We can express the relationship between weights and quality as:
\begin{equation}
w_i = 
\begin{cases}
1.0, & \text{highest quality} \\
\alpha, & \text{lower quality} \quad (0 < \alpha < 1)
\end{cases}
\end{equation}
This weighting scheme guides the model to favor high-quality responses while still learning from a diverse range of instruction-response pairs. We labeled our dataset, as described in Section \ref{Post-training-Data}, and assigned weights to the instruction-response pairs based on predefined rules:
\begin{equation}
w_i = 
\begin{cases}
1.0, & \text{high quality} \\
0.7, & \text{medium quality} \\
0.5, & \text{low quality}
\end{cases}
\end{equation}
where: \\
\noindent \textbf{high quality} - instructions and dialogues manually written by annotators, the OpenHermes-2.5 \cite{OpenHermes2.5} and orca-math-word-problems-200k \cite{mitra2024orcamath} datasets.\\
\noindent \textbf{medium quality} - generated instructions based on pre-training data, which have been manually verified and corrected.\\
\noindent \textbf{low quality} - generated instructions based on pre-training data without manual verification.

We include low-quality instructions with reduced weight (0.5) for several reasons. First, they significantly increase training data volume and diversity, exposing the model to a broader range of linguistic patterns and edge cases that may not be well-represented in smaller high-quality datasets. Second, the weighted loss mechanism ensures these samples contribute less to the optimization objective while still providing useful training signal—effectively allowing the model to learn from imperfect data without compromising performance on high-quality benchmarks. Third, completely discarding these automatically generated samples would waste potentially valuable information, as many contain correct task structures and partially useful content despite quality limitations. This approach of learning from mixed-quality data with appropriate weighting has been shown to improve model robustness and generalization \cite{NEURIPS2023_ac662d74,chen2024alpagasustrainingbetteralpaca}.

Supervised Fine-Tuning (SFT) methods are designed to adapt a pre-trained language model $\pi_0$ into a fine-tuned model $\pi_{\text{SFT}}$ using a high-quality instruction dataset $\mathcal{D}$ and supervised learning. We use $\pi(y|x)$ to represent the probability of generating response $y$ given instruction $x$ in the dataset $\mathcal{D}$. The objective of SFT can be expressed as a maximum likelihood estimate (MLE):
\begin{equation}
J_{\text{SFT}} = \mathbb{E}_{(x,y)\sim\mathcal{D}}[\log \pi_{\text{SFT}}(y|x)]
\end{equation}
To ensure optimal fine-tuning performance, SFT requires the instruction dataset $\mathcal{D}$ to be of the highest possible quality, as it treats all training data uniformly \cite{NEURIPS2023_ac662d74,chen2024alpagasustrainingbetteralpaca}. However, assembling a sufficiently large and high-quality dataset can be both time-consuming and financially expensive.

In practice, the quality of available instructions often varies. It is possible that valuable and informative instructions may have lower quality than desired. To leverage the potential of such mixed-quality data, we introduce the weighted instruction cross-entropy loss, which guides the learning process to prioritize more preferred answers while still allowing the model to learn valuable insights from lower-quality instructions.

The standard Weighted Cross-Entropy Loss \cite{King_Zeng_2001}, originating from the weighted exogenous sampling maximum-likelihood estimator, is frequently used in multi-class classification problems \cite{Wu2024MisclassificationguidedLU}. It is commonly employed, for instance, to address imbalanced class distributions \cite{RezaeiDastjerdehei2020AddressingII}. 
We can formulate standard Weighted Cross-Entropy Loss as follows:
\begin{equation}\label{standard-wcel}
l(o_i, y_i) = - \sum_{c=1}^C w_c \cdot y_{i,c} \cdot \log p_{i,c}
\end{equation}
where $C$ is the number of classes, $\mathbf{y}_i = (y_{i,1}, \dots, y_{i,C}) \in \{0, 1\}^C$ is the one-hot encoding of the ground truth label for sample $x_i$, and $y_{i,c} = 1$ indicates that $x_i$ belongs to class $c$. Meanwhile, $\mathbf{p}_i = (p_{i,1}, \dots, p_{i,C}) \in \mathbb{R}_{+}^C$ represents the predicted probability vector for sample $x_i$ across $C$ classes. In multi-class classification problems using deep neural networks, $p_i$ corresponds to the softmax values of the logits for each class produced by the last layer of the network. Specifically, $p_{i,c} = \frac{\exp(o_{i,c})}{\sum_{j=1}^C \exp(o_{i,j})}$, where $o_{i,c}$ is the logit for class $c$ for sample $x_i$.

To integrate fine-grained weights from the dataset $\mathcal{D}_w$, we modify Eq.~\ref{standard-wcel} as follows:
\begin{equation}\label{rewarded-wcel}
l(o_i, y_i) = -w_i \cdot \sum_{c=1}^C y_{i,c} \cdot \log p_{i,c}
\end{equation}
where $w_i$ represents the weight assigned to the instruction-response pair $(x_i, y_i)$. This learning objective provides a flexible framework for fine-tuning language models, offering more granular control over the importance of each instruction during training. It can capture subtle differences in data quality while maintaining computational efficiency.

\subsection{Training Hyperparameters}

\begin{figure}[h]
\centering
\includegraphics{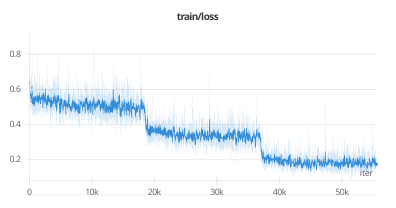}
\caption{Training loss over the training iterations for the instruction model.}
\label{fig:sft-training-loss}
\end{figure}

\begin{figure}[h]
\centering
\includegraphics{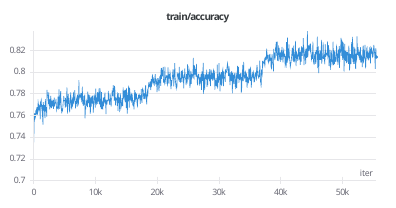}
\caption{Training accuracy over the training iterations for the instruction model.}
\label{fig:sft-training-acc}
\end{figure}

\begin{figure}[h]
\centering
\includegraphics{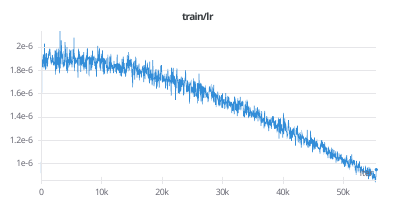}
\caption{Adaptive learning rate over the training iterations for the instruction model.}
\label{fig:sft-training-lr}
\end{figure}

We applied the AdamW optimizer, using $\beta_1 = 0.9$, $\beta_2 = 0.95$, and a weight decay of 0.05. The adaptive learning rate followed a cosine decay, starting at 7e-6 and tapering down to 6e-7, with 50 warmup iterations. The training process spanned a total of 55,440 iterations. Our setup used a global batch size of 128, made up of local batches with a size of 1. Gradient clipping was enforced with a threshold of 1.0, and the model was trained in mixed precision using bfloat16. We trained the model for 3 epochs with a maximum context length of 4,096, processing a total of 2.1 billion tokens. 
The training loss, accuracy, and adaptive learning rate over the training iterations for the instruction model are presented in Figures \ref{fig:sft-training-loss}, \ref{fig:sft-training-acc}, and \ref{fig:sft-training-lr}.

\subsection{Efficient Implementation}
For our training needs, we utilized the ALLaMo framework \cite{allamo}, developed by a co-author of the Bielik model to optimize training throughput. This framework allowed us to maximize the computational resources of the supercomputer, enabling faster calculations and reducing overall training time. ALLaMo achieves high efficiency through numerous optimizations at the dataloader, model, and training process levels, along with a strong reliance on torch.compile in conjunction with an efficient attention implementation using PyTorch SDPA and the PyTorch Fused AdamW optimizer \cite{Ansel_PyTorch_2_Faster_2024}. A significant advantage of ALLaMo is its primary reliance on PyTorch, without dependencies on other popular frameworks or libraries, which allows for better optimization and easier implementation of functionalities not available in other frameworks. For post-training, we implemented the weighted instruction cross-entropy loss and adaptive learning rate strategies, detailed in sections \ref{WICEL} and \ref{ALR}. These improvements enabled us to efficiently conduct numerous experiments and successfully complete the final model training. During the base model training, we utilized 256 NVIDIA GH200 GPUs, achieving a throughput of over 9,200 tokens per GPU per second.

To validate the performance of the ALLaMo framework, we conducted a comparison with the implementation used in training the TinyLlama model \cite{zhang2024tinyllamaopensourcesmalllanguage}. The authors of this model introduced numerous improvements to accelerate training, including FlashAttention-2 \cite{dao2023flashattention2fasterattentionbetter}, fused LayerNorm, fused SwiGLU, fused Cross-Entropy Loss, and fused Rotary Positional Embeddings. The experiment was carried out on A100 40GB GPUs in 8x and 16x A100 configurations, using a model with parameters identical to the TinyLlama 1.1B model. When using ALLaMo, it was possible to increase the local batch size from 8 to 9, which further enhanced training throughput. Table \ref{tab:performance-comparison} illustrates the performance differences between the TinyLlama implementation and the ALLaMo framework.

\begin{table*}[h]
\centering
\begin{tabular}{llll}
\toprule
Framework & Configuration & Total Batch Size & Throughput \\
\midrule
TinyLlama & 8xA100 40GB & 2,097,152 tokens & 24,390 tokens/GPU/sec \\
ALLaMo & 8xA100 40GB & 2,097,152 tokens & 26,150 tokens/GPU/sec (+7.2\%) \\
ALLaMo & 8xA100 40GB & 2,359,296 tokens & 26,550 tokens/GPU/sec (+8.8\%) \\
TinyLlama & 16xA100 40GB & 2,097,152 tokens & 24,000 tokens/GPU/sec $^1$ \\
ALLaMo & 16xA100 40GB & 2,097,152 tokens & 25,850 tokens/GPU/sec (+7.7\%) \\
ALLaMo & 16xA100 40GB & 2,359,296 tokens & 26,000 tokens/GPU/sec (+8.3\%) \\
\bottomrule
\multicolumn{4}{l}{$^1$ \small The value reported by the authors of the model.} \\
\end{tabular}
\caption{A comparison of the training performance between the TinyLlama implementation and the ALLaMo framework.}
\label{tab:performance-comparison}
\end{table*}

\section{Evaluations}
\subsection{Open PL LLM Leaderboard} \label{Open-PL-LLM-Leaderboard}

The Open PL LLM Leaderboard, based on the Open LLM Leaderboard v1 \cite{open-llm-leaderboard-v1}, evaluates models on various NLP tasks, including: sentiment analysis, categorization, and text classification, but does not test their conversational capabilities \cite{open-pl-llm-leaderboard}. The leaderboard utilizes the lm-evaluation-harness framework for model evaluation \cite{eval-harness}.

\paragraph{Tasks:}
\begin{itemize}
    \item \textbf{polemo2:} Sentiment analysis of online consumer reviews across four domains (medicine, hotels, products, university) with four-class labeling (positive, negative, neutral, ambiguous) \cite{kocon-etal-2019-multi}; metric: accuracy.
    \item \textbf{klej-ner:} Named entity recognition in sentences containing single-type entities, classifying into six categories (no entity, place, person, organization, time, geographical name) \cite{rybak-etal-2020-klej}; metric: accuracy.
    \item \textbf{8tags:} Topic classification of social media headlines into eight categories (film, history, food, medicine, motorization, work, sport, technology) \cite{dadas-etal-2020-evaluation}; metric: accuracy.
    \item \textbf{belebele:} Machine reading comprehension for question answering \cite{bandarkar-etal-2024-belebele}; metric: accuracy.
    \item \textbf{dyk:} Question answering based on human-annotated pairs from Wikipedia's "Did You Know" section \cite{marcinczuk2013open}; metric: binary F1.
    \item \textbf{ppc:} Text similarity assessment using manually labeled sentence pairs (exact paraphrases, close paraphrases, non-paraphrases) \cite{9945218}; metric: accuracy.
    \item \textbf{psc:} Summarization of news articles \cite{ogro:kop:14:lrec}; metric: binary F1.
    \item \textbf{cbd:} Text classification for cyberbullying and hate-speech detection \cite{ptaszynski2023expert}; metric: macro F1.
    \item \textbf{polqa:} Open-domain question answering from the "Jeden z dziesięciu" TV show, with and without context (abstractive QA/RAG) \cite{rybak-etal-2024-polqa-polish}; metric: accuracy, levenshtein.
    \item \textbf{poquad:} Context-based extractive question answering (QA/RAG) \cite{tuora2023poquad}; metric: levenshtein.
\end{itemize}

Most of the tasks are multiple-choice tests, which means that the model chooses the correct answer from a set of options.
They are implemented as two types of tests:
\begin{itemize}
    \item \textbf{Loglikelihood:} We choose the highest probability token from the given set, e.g., ABCD. These tests are suitable for base models.
    \item \textbf{Generate:} Model generates answer freely.
\end{itemize}

All tasks are evaluated in both 0-shot and 5-shot settings.

\paragraph{Evaluation scores:}
\begin{itemize}
    \item \textbf{All tasks:} The average score across all tasks, normalized by baseline scores.
    \item \textbf{Reranking:} The score of the polqa\_reranking\_mc task, which is based on the polqa dataset. This task evaluates the model's ability to determine whether a given context is relevant to a question (binary relevance classification). This capability is essential in Retrieval-Augmented Generation (RAG) systems for reranking retrieved documents based on their relevance to a query.
    \item \textbf{Reader (Generator):} The average score of open-book question-answering tasks (polqa and poquad), which evaluate the model's ability to generate or extract answers from provided context. These tasks directly measure performance in the reader component of RAG systems, where the model must comprehend retrieved documents and produce accurate answers.
    \item \textbf{Perplexity:} A bonus metric that does not correlate with other scores and should not be used for direct model comparison (lower is better).
\end{itemize}

As of April 3, 2024, Table \ref{tab:model-comparison} presents the current scores of both pretrained and continuously pretrained models, as evaluated on the Open PL LLM Leaderboard in a 5-shot setting.

The Bielik 7B v0.1 base model demonstrates strong performance in RAG-specific tasks. Among base models, Bielik 7B v0.1 achieves the highest RAG Reader score (88.39), representing a notable improvement of approximately 3 percentage points over Mistral-7B-v0.1 (85.39). However, it is important to note that this specialized performance comes with a trade-off: Bielik's overall average score across all tasks (29.38) is slightly lower than Mistral-7B-v0.1 (30.67). This reflects our base model's optimization focus on Polish language understanding and RAG capabilities, which resulted in particularly strong performance on context-based question answering tasks at the potential expense of broader task coverage. 

Similarly, the instruction-tuned Bielik-7B-Instruct-v0.1 achieves exceptional RAG Reader performance (86.00) compared to Mistral-7B-Instruct-v0.1 (73.68), representing nearly a 9 percentage point improvement, though Bielik maintains a higher overall score (39.28 vs 26.42) across all tasks in this comparison. In our subjective evaluations of conversational abilities, our models outperformed others that had higher average scores. The results presented in Table \ref{tab:model-comparison} were obtained without employing instruction templates for the instructional models, treating them instead as base models. This approach may have skewed the results, as instructional models are specifically optimized to follow particular instructions.

\begin{table*}[t]
\centering
\small
\begin{tabular}{lrrrr}
\toprule
\textbf{Model} & \textbf{All tasks} & \textbf{RAG} & \textbf{RAG} & \textbf{Perplexity} \\
& & \textbf{Reranking} & \textbf{Reader} & \\
\midrule
\multicolumn{5}{l}{\textbf{7B parameters models:}} \\
berkeley-nest/Starling-LM-7B-alpha & \textbf{47.46} & \textbf{75.73} & 82.86 & 1438.04 \\
openchat/openchat-3.5-0106 & 47.32 & 74.71 & 83.60 & 1106.56 \\
Nexusflow/Starling-LM-7B-beta & 45.69 & 74.58 & 81.22 & 1161.54 \\
openchat/openchat-3.5-1210 & 44.17 & 71.76 & 82.15 & 1923.83 \\
teknium/OpenHermes-2.5-Mistral-7B & 42.64 & 70.63 & 80.25 & 1463.00 \\
mistralai/Mistral-7B-Instruct-v0.2 & 40.29 & 72.58 & 79.39 & 2088.08 \\
\underline{Bielik-7B-Instruct-v0.1} & 39.28 & 61.89 & \textbf{86.00} & 277.92 \\
internlm/internlm2-chat-7b & 37.64 & 72.29 & 71.17 & 3892.50 \\
internlm/internlm2-chat-7b-sft & 36.97 & 73.22 & 69.96 & 4269.63 \\
HuggingFaceH4/zephyr-7b-alpha & 33.97 & 71.47 & 73.35 & 4464.45 \\
HuggingFaceH4/zephyr-7b-beta & 33.15 & 71.65 & 71.27 & 3613.14 \\
szymonrucinski/Curie-7B-v1 & 26.72 & 55.58 & 85.19 & 389.17 \\
mistralai/Mistral-7B-Instruct-v0.1 & 26.42 & 56.35 & 73.68 & 6909.94 \\
meta-llama/Llama-2-7b-chat-hf & 21.04 & 54.65 & 72.93 & 4018.74 \\
Voicelab/trurl-2-7b & 18.85 & 60.67 & 77.19 & 1098.88 \\
Baseline (majority class) & 0.00 & 53.36 & - & - \\
\midrule
\multicolumn{5}{l}{\textbf{Models with different sizes:}} \\
upstage/SOLAR-10.7B-Instruct-v1.0 (10.7B) & 46.07 & \textbf{76.93} & 82.86 & 789.58 \\
Voicelab/trurl-2-13b-academic (13B) & 29.45 & 68.19 & 79.88 & 733.91 \\
Azurro/APT3-1B-Instruct-v1 (1B) & -13.80 & 52.11 & 12.23 & 739.09 \\
\midrule
\multicolumn{5}{l}{\textbf{7B parameters pretrained and continuously pretrained models:}} \\
alpindale/Mistral-7B-v0.2-hf & \textbf{33.05} & 60.23 & 85.21 & 932.60 \\
internlm/internlm2-7b & 33.03 & \textbf{69.39} & 73.63 & 5498.23 \\
mistralai/Mistral-7B-v0.1 & 30.67 & 60.35 & 85.39 & 857.32 \\
\underline{Bielik-7B-v0.1} & 29.38 & 62.13 & \textbf{88.39} & 123.31 \\
internlm/internlm2-base-7b & 20.68 & 52.39 & 69.85 & 3110.92 \\
meta-llama/Llama-2-7b-hf & 12.73 & 54.02 & 77.92 & 850.45 \\
OPI-PG/Qra-7b & 11.13 & 54.40 & 75.25 & 203.36 \\
\bottomrule
\end{tabular}
\caption{Detailed comparison among Bielik 7B v0.1 and other representative open-source models}
\label{tab:model-comparison}
\end{table*}


\subsection{Polish MT-Bench}

MT-bench \cite{zheng2023judging} is a tool designed to test the ability of language models (LLMs) to conduct two-step conversations and follow instructions. It covers typical use cases and focuses on challenging questions to differentiate the capabilities of various models. Eight main categories of user queries were identified, which were used to construct MT-bench:

\begin{itemize}
    \item writing
    \item role-playing
    \item information extraction
    \item reasoning
    \item mathematics
    \item coding
    \item knowledge / hard sciences / stem
    \item knowledge / humanities / social sciences
\end{itemize}

For each category, two-step questions were manually developed.

The evaluation of responses is performed by a metamodel. In the case of MT-Bench, this is the GPT-4 model. By using a metamodel, we can verify responses from open-ended questions, e.g., write an article about hybrid cars. The model evaluates the content of the response, the quality of facts used, creativity, etc.

The Polish MT-Bench \cite{rkinas2024mt-bench-pl} has been completely polonized. Each task was first machine-translated and then verified. Additionally, we introduced Polish accents, e.g., instead of describing a vacation in Hawaii, we suggested the location - Masuria. In our language version, many changes were introduced to transfer the test into Polish linguistic realities.

\begin{table*}[t]
\centering
\begin{tabular}{lrrr}
\toprule
\textbf{Model} & \textbf{pl\_score} & \textbf{responses\_pl} & \textbf{Average Score} \\
\midrule
Mixtral-8x7b & \textbf{7.64} & 1.00 & \textbf{7.64} \\
Mistral-Nemo-Instruct-2407 & 7.37 & 1.00 & 7.37 \\
openchat-3.5-0106-gemma & 6.51 & 0.96 & 6.81 \\
Meta-Llama-3.1-8B-Instruct & 6.24 & 1.00 & 6.24 \\
Starling-LM-7B-alpha & 6.05 & 0.93 & 6.49 \\
openchat-3.5-0106 & 6.03 & 0.94 & 6.39 \\
Mistral-7B-Instruct-v0.3 & 5.75 & 0.98 & 5.82 \\
\underline{Bielik-7B-Instruct-v0.1} & 5.40 & 0.89 & 6.08 \\
dolphin-2.9.1-llama-3-8b & 5.24 & 0.89 & 5.86 \\
Polka-Mistral-7B-SFT & 4.43 & 0.98 & 4.52 \\
trurl-2-7b & 2.75 & 0.99 & 2.76 \\
Mistral-7B-Instruct-v0.2 & 2.05 & 0.31 & 6.56 \\
\bottomrule
\end{tabular}
\caption{Polish MT-Bench results for various language models}
\label{tab:pl-mt-bench-avg}
\end{table*}

Table \ref{tab:pl-mt-bench-avg} presents the results of the Polish MT-Bench evaluation for various language models. The table shows three key metrics: the Polish score (pl\_score), the proportion of responses in Polish (responses\_pl), and the average score. The Bielik 7B v0.1 model has a pl\_score of 5.40, demonstrating competitive performance among larger models.

\begin{table*}[t]
\centering
\tiny
\begin{tabular}{lrrrrrrrr}
\toprule
\textbf{Model} & \textbf{Coding} & \textbf{Extraction} & \textbf{Humanities} & \textbf{Mathematics} & \textbf{Reasoning} & \textbf{Role-playing} & \textbf{Stem} & \textbf{Writing} \\
\midrule
Mixtral-8x7b & 5.20 & 8.15 & 9.45 & 5.65 & 5.80 & \textbf{8.95} & 8.55 & \textbf{9.35} \\
Mistral-Nemo-Instruct-2407 & \textbf{5.85} & 8.95 & \textbf{9.50} & \textbf{6.70} & 5.80 & 7.45 & 8.30 & 6.40 \\
openchat-3.5-0106-gemma & 5.35 & 6.90 & 8.80 & 4.55 & 5.40 & 7.97 & 8.47 & 7.05 \\
Meta-Llama-3.1-8B-Instruct & 4.60 & \textbf{9.10} & 8.82 & 5.30 & 2.50 & 5.60 & 6.30 & 7.70 \\
Starling-LM-7B-alpha & 4.75 & 7.35 & 8.50 & 4.15 & 3.90 & 6.90 & \textbf{8.85} & 7.55 \\
openchat-3.5-0106 & 5.05 & 6.90 & 9.30 & 3.80 & 3.90 & 6.00 & 8.40 & 7.75 \\
Mistral-7B-Instruct-v0.3 & 4.30 & 7.30 & 6.75 & 2.35 & 3.80 & 7.25 & 7.45 & 7.35 \\
\underline{Bielik-7B-Instruct-v0.1} & 3.00 & 4.35 & 8.47 & 4.10 & \textbf{6.15} & 7.83 & 6.90 & 7.85 \\
dolphin-2.9.1-llama-3-8b & 4.60 & 6.15 & 8.80 & 4.80 & 3.30 & 7.40 & 6.35 & 5.50 \\
Polka-Mistral-7B-SFT & 2.95 & 5.25 & 5.60 & 2.95 & 2.45 & 4.90 & 6.80 & 5.25 \\
trurl-2-7b & 1.80 & 3.50 & 3.95 & 1.70 & 2.05 & 3.30 & 2.65 & 3.15 \\
Mistral-7B-Instruct-v0.2 & 4.25 & 7.40 & 8.40 & 3.20 & 5.00 & - & - & - \\
\bottomrule
\end{tabular}
\caption{Polish MT-Bench results for various language models by category}
\label{tab:pl-mt-bench}
\end{table*}

Table \ref{tab:pl-mt-bench} provides a more detailed breakdown of the Polish MT-Bench results, showing scores across eight different categories for each model. The Bielik 7B v0.1 model shows competitive performance in several categories, notably excelling in Reasoning (6.15) and Role-playing (7.83). These results demonstrate the model's versatility across various tasks, despite its smaller size compared to some of the top-performing models.

\FloatBarrier

\subsection{Bias, Toxicity and Misinformation}

Language models have been shown to reproduce and amplify biases present in the training data, and can generate toxic or offensive content. Since our training dataset contains a large proportion of data from the web, Bielik-7B-v0.1 may produce factually incorrect output and should not be relied upon for producing accurate information. Despite significant efforts to clean the training data, it is still possible for this model to generate lewd, false, biased, or otherwise offensive content.

\section{Model Quantization}
In our work on the Bielik 7B v0.1 model, our primary objective was to create quantized versions that could be accessible to users with limited computational resources. This effort was driven by a vision to democratize advanced language models and make them available to those who do not have access to powerful computing infrastructure. By optimizing our model for low-resource environments, we aimed to facilitate deployment on various devices, including edge devices such as mobile phones and embedded systems.

To achieve this, we developed and delivered several quantized versions of Bielik 7B v0.1, including GGUF (GPT -
Generated Unified Format)\footnote{\url{https://github.com/ggerganov/ggml}}, HQQ (Half-Quadratic Quantization) \cite{badri2023hqq}, AWQ (Activation-aware Weight Quantization) \cite{lin2023awq}, MLX (Apple MLX Framework) \cite{mlx2023}, EXL2 (ExLlamaV2)\footnote{\url{https://github.com/turboderp/exllamav2}}, GPTQ (Accurate Post-Training Quantization for Generative Pre-trained Transformers) \cite{frantar-gptq}, and IQ2\_XXS (GGUF IQ)\footnote{\url{https://github.com/ggerganov/llama.cpp}}. Each quantization technique offered different trade-offs in terms of performance, memory usage, and computational requirements, allowing for flexibility depending on the intended use case and hardware capabilities. The IQ2\_XXS version, in particular, was specifically designed for edge devices, with a bit-per-weight quantization of 2.06 bpw, providing an efficient solution for deployment on resource-constrained platforms such as mobile phones.

\subsection{Calibration and Evaluation of Quantized Models}

In addition to the standard quantization process, we created calibrated versions of the imatrix (Importance Matrix) GGUF model. Calibration plays a crucial role in minimizing performance degradation, which is often a concern during quantization. To support this process, we developed a multilingual (Polish-English) calibration dataset with a specific emphasis on the Polish language. This multilingual approach aimed to improve the model's generalization capabilities across languages while ensuring high fidelity in its Polish-language outputs.

To assess the impact of calibration, we conducted a thorough comparison between the uncalibrated and calibrated versions of the model for the Polish language. Our evaluation metrics focused on both the accuracy of language understanding and the quality of generated text. The results showed that the calibration process improved the model's performance, particularly in language-specific contexts where nuances and subtleties are crucial.

Across all quantization schemes examined (Q8\_0, Q6\_K, Q5\_K\_M, Q4\_K\_M, Q3\_K\_M, Q2\_K) (see table \ref{tab:polish-model-quantization}), models quantized with imatrix consistently outperform their counterparts without imatrix quantization. This is evident through multiple evaluation metrics, indicating that imatrix quantization effectively preserves model quality even at lower bit-widths. The KLD values are consistently lower for imatrix-quantized models, indicating a closer alignment of the probability distributions between the quantized and the original FP16 models. Imatrix quantization results in Mean $\Delta$p values closer to zero, indicating less degradation in the model's ability to predict the correct token. At Q3\_K\_M, the Mean $\Delta$p improves from -0.9160 without imatrix to -0.3860 with imatrix. The advantages of imatrix quantization become more pronounced at lower bit-width quantization levels. The reduction in performance metrics such as PPL, KLD, Mean $\Delta$p, and RMS $\Delta$p is more significant when comparing imatrix and non-imatrix models at Q2\_K and Q3\_K\_M levels, demonstrating imatrix's effectiveness in mitigating the adverse effects of aggressive quantization.

The application of imatrix quantization to the Polish language model leads to significant improvements in maintaining model quality across various quantization levels. These findings support the adoption of imatrix quantization as an effective technique for compressing language models without substantially compromising their performance.

\begin{table}[!htbp]
\centering
\resizebox{\columnwidth}{!}{%
\begin{tabular}{lcccccccc}
\toprule
\textbf{Quant.} & \textbf{imatrix} & \textbf{Size} & \textbf{PPL} & \textbf{$\Delta$PPL} & \textbf{KLD} & \textbf{Mean $\Delta$p} & \textbf{RMS $\Delta$p} & \textbf{Same top p} \\
 & & \textbf{[GiB]} & & & & & & \textbf{[\%]} \\
\midrule
FP16 & - & 13.49 & 3.9393 & - & - & - & - & - \\
Q8\_0 & No & 7.17 & 3.9422 & 0.0029 & 0.0010 & -0.0070 & 0.9800 & 98.6890 \\
Q8\_0 & Yes & 7.17 & 3.9422 & 0.0029 & 0.0010 & -0.0070 & 0.9800 & 98.6890 \\
Q6\_K & No & 5.53 & 3.9450 & 0.0057 & 0.0051 & -0.0420 & 2.1850 & 97.2410 \\
\textbf{Q6\_K} & \textbf{Yes} & 5.53 & \textbf{3.9406} & 0.0013 & 0.0037 & -0.0030 & 1.8490 & 97.6130 \\
Q5\_K\_M & No & 4.78 & 3.9520 & 0.0127 & 0.0106 & -0.0680 & 3.1320 & 96.0510 \\
\textbf{Q5\_K\_M} & \textbf{Yes} & 4.78 & \textbf{3.9473} & 0.0080 & 0.0086 & -0.0250 & 2.8320 & 96.4670 \\
Q4\_K\_M & No & 4.07 & 3.9876 & 0.0483 & 0.0286 & -0.2690 & 5.1300 & 93.6550 \\
\textbf{Q4\_K\_M} & \textbf{Yes} & 4.07 & \textbf{3.9727} & 0.0333 & 0.0220 & -0.1440 & 4.4880 & 94.4700 \\
Q3\_K\_M & No & 3.28 & 4.0915 & 0.1522 & 0.0826 & -0.9160 & 8.6880 & 89.5780 \\
\textbf{Q3\_K\_M} & \textbf{Yes} & 3.28 & \textbf{4.0458} & 0.1065 & 0.0683 & -0.3860 & 7.8390 & 90.6290 \\
Q2\_K & No & 2.53 & 4.7045 & 0.7652 & 0.2852 & -3.8050 & 16.3760 & 81.1100 \\
\textbf{Q2\_K} & \textbf{Yes} & 2.53 & \textbf{4.3522} & 0.4128 & 0.1939 & -1.8980 & 13.4190 & 84.5580 \\
\bottomrule
\end{tabular}%
}
\caption{Comparison of quantization results for the Bielik 7B v0.1 model using imatrix: PPL - Perplexity, $\Delta$PPL - change in perplexity, KLD - Kullback-Leibler Divergence, Mean $\Delta$p - mean change in correct token probability, RMS $\Delta$p - root mean square of change in token probabilities, Same top p - the percentage of instances where the quantized model and the FP16 model assign the highest probability to the same token.}
\label{tab:polish-model-quantization}
\end{table}

\section{Conclusion}

In this paper, we introduced Bielik 7B v0.1, a language model specifically trained for the Polish language. We demonstrated that it is possible to significantly enhance the linguistic capabilities of an already trained model by fine-tuning it on texts exclusively in that language. Without changing the tokenizer, we achieved a high quality of responses generated by the model, which resembled texts written by native Polish speakers. Furthermore, the model performed well in various tasks, opening up intriguing possibilities for its further development.

\begin{acknowledgements}

We gratefully acknowledge Poland's high-performance Infrastructure PLGrid ACK Cyfronet AGH for providing computer facilities and support within computational grant no PLG/2024/016951.

The model could not have been created without the commitment and work of the entire SpeakLeash team, whose contribution is invaluable. Thanks to the hard work of many individuals, it was possible to gather a large amount of content in Polish and establish collaboration between the open-science SpeakLeash project and the HPC center: ACK Cyfronet AGH. Individuals who contributed to the creation of the model through their commitment to the open-science SpeakLeash project: Sebastian Kondracki, Szymon Mazurek, Maria Filipkowska, Paweł Kiszczak, Igor Ciuciura, Jacek Chwiła, Szymon Baczyński, Grzegorz Urbanowicz, Paweł Cyrta, Jan Maria Kowalski, Karol Jezierski, Kamil Nonckiewicz, Izabela Babis, Nina Babis, Waldemar Boszko, and many other wonderful researchers and enthusiasts of the AI world.

\end{acknowledgements}

\bibliographystyle{cs-agh}
\bibliography{bibliography}

@article{jiang2023mistral7b,
 author = {Albert Q. Jiang and Alexandre Sablayrolles and Arthur Mensch and Chris Bamford and Devendra Singh Chaplot and Diego de las Casas and Florian Bressand and Gianna Lengyel and Guillaume Lample and Lucile Saulnier and Lélio Renard Lavaud and Marie-Anne Lachaux and Pierre Stock and Teven Le Scao and Thibaut Lavril and Thomas Wang and Timothée Lacroix and William El Sayed},
 journal = {ArXiv preprint},
 title = {Mistral 7B},
 url = {https://arxiv.org/abs/2310.06825},
 volume = {abs/2310.06825},
 year = {2023}
}

@inproceedings{Vaswani2017AttentionIA,
 author = {Ashish Vaswani and
Noam Shazeer and
Niki Parmar and
Jakob Uszkoreit and
Llion Jones and
Aidan N. Gomez and
Lukasz Kaiser and
Illia Polosukhin},
 bibsource = {dblp computer science bibliography, https://dblp.org},
 booktitle = {Advances in Neural Information Processing Systems 30: Annual Conference
on Neural Information Processing Systems 2017, December 4-9, 2017,
Long Beach, CA, {USA}},
 editor = {Isabelle Guyon and
Ulrike von Luxburg and
Samy Bengio and
Hanna M. Wallach and
Rob Fergus and
S. V. N. Vishwanathan and
Roman Garnett},
 pages = {5998--6008},
 title = {Attention is All you Need},
 url = {https://proceedings.neurips.cc/paper/2017/hash/3f5ee243547dee91fbd053c1c4a845aa-Abstract.html},
 year = {2017}
}

@inproceedings{ainslie-etal-2023-gqa,
 address = {Singapore},
 author = {Ainslie, Joshua  and
Lee-Thorp, James  and
de Jong, Michiel  and
Zemlyanskiy, Yury  and
Lebron, Federico  and
Sanghai, Sumit},
 booktitle = {Proceedings of the 2023 Conference on Empirical Methods in Natural Language Processing},
 doi = {10.18653/v1/2023.emnlp-main.298},
 editor = {Bouamor, Houda  and
Pino, Juan  and
Bali, Kalika},
 pages = {4895--4901},
 publisher = {Association for Computational Linguistics},
 title = {{GQA}: Training Generalized Multi-Query Transformer Models from Multi-Head Checkpoints},
 url = {https://aclanthology.org/2023.emnlp-main.298},
 year = {2023}
}

@article{Child2019GeneratingLS,
 author = {Rewon Child and Scott Gray and Alec Radford and Ilya Sutskever},
 journal = {ArXiv preprint},
 title = {Generating Long Sequences with Sparse Transformers},
 url = {https://arxiv.org/abs/1904.10509},
 volume = {abs/1904.10509},
 year = {2019}
}

@article{Beltagy2020LongformerTL,
 author = {Iz Beltagy and Matthew E. Peters and Arman Cohan},
 journal = {ArXiv preprint},
 title = {Longformer: The Long-Document Transformer},
 url = {https://arxiv.org/abs/2004.05150},
 volume = {abs/2004.05150},
 year = {2020}
}

@inproceedings{Dauphin2016LanguageMW,
 author = {Yann N. Dauphin and
Angela Fan and
Michael Auli and
David Grangier},
 bibsource = {dblp computer science bibliography, https://dblp.org},
 booktitle = {Proceedings of the 34th International Conference on Machine Learning,
{ICML} 2017, Sydney, NSW, Australia, 6-11 August 2017},
 editor = {Doina Precup and
Yee Whye Teh},
 pages = {933--941},
 publisher = {{PMLR}},
 series = {Proceedings of Machine Learning Research},
 title = {Language Modeling with Gated Convolutional Networks},
 url = {http://proceedings.mlr.press/v70/dauphin17a.html},
 volume = {70},
 year = {2017}
}

@article{shazeer2020gluvariantsimprovetransformer,
 author = {Noam Shazeer},
 journal = {ArXiv preprint},
 title = {GLU Variants Improve Transformer},
 url = {https://arxiv.org/abs/2002.05202},
 volume = {abs/2002.05202},
 year = {2020}
}

@article{SU2024127063,
 author = {Jianlin Su and Murtadha Ahmed and Yu Lu and Shengfeng Pan and Wen Bo and Yunfeng Liu},
 doi = {https://doi.org/10.1016/j.neucom.2023.127063},
 issn = {0925-2312},
 journal = {Neurocomputing},
 pages = {127063},
 title = {RoFormer: Enhanced transformer with Rotary Position Embedding},
 url = {https://www.sciencedirect.com/science/article/pii/S0925231223011864},
 volume = {568},
 year = {2024}
}

@inproceedings{10.5555/3666122.3668105,
 author = {Zixuan Jiang and
Jiaqi Gu and
Hanqing Zhu and
David Z. Pan},
 bibsource = {dblp computer science bibliography, https://dblp.org},
 booktitle = {Advances in Neural Information Processing Systems 36: Annual Conference
on Neural Information Processing Systems 2023, NeurIPS 2023, New Orleans,
LA, USA, December 10 - 16, 2023},
 editor = {Alice Oh and
Tristan Naumann and
Amir Globerson and
Kate Saenko and
Moritz Hardt and
Sergey Levine},
 title = {Pre-RMSNorm and Pre-CRMSNorm Transformers: Equivalent and Efficient
Pre-LN Transformers},
 url = {http://papers.nips.cc/paper\_files/paper/2023/hash/8f1bacee31caf990a4f08d84f0ccb322-Abstract-Conference.html},
 year = {2023}
}

@misc{AzurroAPT3Base1B,
 author = {Ociepa, Krzysztof and {Azurro Team}},
 note = {Accessed: 2024-09-30},
 title = {Introducing APT3-1B-Base: Polish Language Model},
 url = {https://azurro.pl/apt3-1b-base-en},
 urldate = {2024-09-30},
 year = {2024}
}

@inproceedings{sennrich2016neuralmachinetranslationrare,
 address = {Berlin, Germany},
 author = {Sennrich, Rico  and
Haddow, Barry  and
Birch, Alexandra},
 booktitle = {Proceedings of the 54th Annual Meeting of the Association for Computational Linguistics (Volume 1: Long Papers)},
 doi = {10.18653/v1/P16-1162},
 editor = {Erk, Katrin  and
Smith, Noah A.},
 pages = {1715--1725},
 publisher = {Association for Computational Linguistics},
 title = {Neural Machine Translation of Rare Words with Subword Units},
 url = {https://aclanthology.org/P16-1162},
 year = {2016}
}

@inproceedings{kudo2018sentencepiecesimplelanguageindependent,
 address = {Brussels, Belgium},
 author = {Kudo, Taku  and
Richardson, John},
 booktitle = {Proceedings of the 2018 Conference on Empirical Methods in Natural Language Processing: System Demonstrations},
 doi = {10.18653/v1/D18-2012},
 editor = {Blanco, Eduardo  and
Lu, Wei},
 pages = {66--71},
 publisher = {Association for Computational Linguistics},
 title = {{S}entence{P}iece: A simple and language independent subword tokenizer and detokenizer for Neural Text Processing},
 url = {https://aclanthology.org/D18-2012},
 year = {2018}
}

@article{imamura2022extendingsubwordingmodelmultilingual,
 author = {Kenji Imamura and Eiichiro Sumita},
 journal = {ArXiv preprint},
 title = {Extending the Subwording Model of Multilingual Pretrained Models for New Languages},
 url = {https://arxiv.org/abs/2211.15965},
 volume = {abs/2211.15965},
 year = {2022}
}

@inproceedings{Li2022OvercomingCF,
 address = {Seattle, United States},
 author = {Li, Dingcheng  and
Chen, Zheng  and
Cho, Eunah  and
Hao, Jie  and
Liu, Xiaohu  and
Xing, Fan  and
Guo, Chenlei  and
Liu, Yang},
 booktitle = {Proceedings of the 2022 Conference of the North American Chapter of the Association for Computational Linguistics: Human Language Technologies},
 doi = {10.18653/v1/2022.naacl-main.398},
 editor = {Carpuat, Marine  and
de Marneffe, Marie-Catherine  and
Meza Ruiz, Ivan Vladimir},
 pages = {5441--5454},
 publisher = {Association for Computational Linguistics},
 title = {Overcoming Catastrophic Forgetting During Domain Adaptation of Seq2seq Language Generation},
 url = {https://aclanthology.org/2022.naacl-main.398},
 year = {2022}
}

@inproceedings{pmlr-v199-ostapenko22a,
 author = {Ostapenko, Oleksiy and Lesort, Timothee and Rodriguez, Pau and Arefin, Md Rifat and Douillard, Arthur and Rish, Irina and Charlin, Laurent},
 booktitle = {Proceedings of The 1st Conference on Lifelong Learning Agents},
 editor = {Chandar, Sarath and Pascanu, Razvan and Precup, Doina},
 pages = {60--91},
 publisher = {PMLR},
 series = {Proceedings of Machine Learning Research},
 title = {Continual Learning with Foundation Models: An Empirical Study of Latent Replay},
 url = {https://proceedings.mlr.press/v199/ostapenko22a.html},
 volume = {199},
 year = {2022}
}

@article{ibrahim2024simplescalablestrategiescontinually,
 author = {Adam Ibrahim and Benjamin Thérien and Kshitij Gupta and Mats L. Richter and Quentin Anthony and Timothée Lesort and Eugene Belilovsky and Irina Rish},
 journal = {ArXiv preprint},
 title = {Simple and Scalable Strategies to Continually Pre-train Large Language Models},
 url = {https://arxiv.org/abs/2403.08763},
 volume = {abs/2403.08763},
 year = {2024}
}

@misc{cerebras2023slimpajama,
 author = {Soboleva, Daria and Al-Khateeb, Faisal and Myers, Robert and Steeves, Jacob R and Hestness, Joel and Dey, Nolan},
 howpublished = {\url{https://www.cerebras.net/blog/slimpajama-a-627b-token-cleaned-and-deduplicated-version-of-redpajama}},
 title = {{SlimPajama: A 627B token cleaned and deduplicated version of RedPajama}},
 url = {https://huggingface.co/datasets/cerebras/SlimPajama-627B},
 year = {2023}
}

@misc{speakleashorg,
 author = {{SpeakLeash Team}},
 note = {Accessed: 2024-09-30},
 title = {SpeakLeash a.k.a Spichlerz!},
 url = {https://www.speakleash.org},
 urldate = {2024-09-30},
 year = {2024}
}

@article{okulska2023stylometrixopensourcemultilingualtool,
 author = {Inez Okulska and Daria Stetsenko and Anna Ko{\l}os and Agnieszka Karli\'{n}ska and Kinga G\l{}\k{a}bi\'{n}ska and Adam Nowakowski},
 journal = {ArXiv preprint},
 title = {StyloMetrix: An Open-Source Multilingual Tool for Representing Stylometric Vectors},
 url = {https://arxiv.org/abs/2309.12810},
 volume = {abs/2309.12810},
 year = {2023}
}

@inproceedings{Loshchilov2017DecoupledWD,
 author = {Ilya Loshchilov and
Frank Hutter},
 bibsource = {dblp computer science bibliography, https://dblp.org},
 booktitle = {7th International Conference on Learning Representations, {ICLR} 2019,
New Orleans, LA, USA, May 6-9, 2019},
 publisher = {OpenReview.net},
 title = {Decoupled Weight Decay Regularization},
 url = {https://openreview.net/forum?id=Bkg6RiCqY7},
 year = {2019}
}

@misc{OpenHermes2.5,
 author = {Teknium},
 howpublished = {\url{https://huggingface.co/datasets/teknium/OpenHermes-2.5}},
 publisher = {HuggingFace},
 title = {OpenHermes 2.5: An Open Dataset of Synthetic Data for Generalist LLM Assistants},
 year = {2023}
}

@misc{mitra2024orcamath,
 archiveprefix = {arXiv},
 author = {Arindam Mitra and Hamed Khanpour and Corby Rosset and Ahmed Awadallah},
 eprint = {2402.14830},
 primaryclass = {cs.CL},
 title = {Orca-Math: Unlocking the potential of SLMs in Grade School Math},
 year = {2024}
}

@inproceedings{NEURIPS2023_ac662d74,
 author = {Chunting Zhou and
Pengfei Liu and
Puxin Xu and
Srinivasan Iyer and
Jiao Sun and
Yuning Mao and
Xuezhe Ma and
Avia Efrat and
Ping Yu and
Lili Yu and
Susan Zhang and
Gargi Ghosh and
Mike Lewis and
Luke Zettlemoyer and
Omer Levy},
 bibsource = {dblp computer science bibliography, https://dblp.org},
 booktitle = {Advances in Neural Information Processing Systems 36: Annual Conference
on Neural Information Processing Systems 2023, NeurIPS 2023, New Orleans,
LA, USA, December 10 - 16, 2023},
 editor = {Alice Oh and
Tristan Naumann and
Amir Globerson and
Kate Saenko and
Moritz Hardt and
Sergey Levine},
 title = {{LIMA:} Less Is More for Alignment},
 url = {http://papers.nips.cc/paper\_files/paper/2023/hash/ac662d74829e4407ce1d126477f4a03a-Abstract-Conference.html},
 year = {2023}
}

@article{Wu2024MisclassificationguidedLU,
 author = {Yanxue Wu and Kai Du and Xian-Jie Wang and Fan Min},
 journal = {Knowl. Inf. Syst.},
 pages = {4685-4720},
 title = {Misclassification-guided loss under the weighted cross-entropy loss framework},
 url = {https://api.semanticscholar.org/CorpusID:269765983},
 volume = {66},
 year = {2024}
}

@article{King_Zeng_2001,
 author = {King, Gary and Zeng, Langche},
 doi = {10.1093/oxfordjournals.pan.a004868},
 journal = {Political Analysis},
 number = {2},
 pages = {137–163},
 title = {Logistic Regression in Rare Events Data},
 volume = {9},
 year = {2001}
}

@inproceedings{pmlr-v162-xu22l,
 author = {Haoran Xu and
Xianyuan Zhan and
Honglei Yin and
Huiling Qin},
 bibsource = {dblp computer science bibliography, https://dblp.org},
 booktitle = {International Conference on Machine Learning, {ICML} 2022, 17-23 July
2022, Baltimore, Maryland, {USA}},
 editor = {Kamalika Chaudhuri and
Stefanie Jegelka and
Le Song and
Csaba Szepesv{\'{a}}ri and
Gang Niu and
Sivan Sabato},
 pages = {24725--24742},
 publisher = {{PMLR}},
 series = {Proceedings of Machine Learning Research},
 title = {Discriminator-Weighted Offline Imitation Learning from Suboptimal
Demonstrations},
 url = {https://proceedings.mlr.press/v162/xu22l.html},
 volume = {162},
 year = {2022}
}

@inproceedings{wang2024openchatadvancingopensourcelanguage,
 author = {Guan Wang and
Sijie Cheng and
Xianyuan Zhan and
Xiangang Li and
Sen Song and
Yang Liu},
 bibsource = {dblp computer science bibliography, https://dblp.org},
 booktitle = {The Twelfth International Conference on Learning Representations,
{ICLR} 2024, Vienna, Austria, May 7-11, 2024},
 publisher = {OpenReview.net},
 title = {OpenChat: Advancing Open-source Language Models with Mixed-Quality
Data},
 url = {https://openreview.net/forum?id=AOJyfhWYHf},
 year = {2024}
}

@article{Granziol2020LearningRA,
 author = {Diego Granziol and Stefan Zohren and Stephen J. Roberts},
 journal = {J. Mach. Learn. Res.},
 pages = {173:1-173:65},
 title = {Learning Rates as a Function of Batch Size: A Random Matrix Theory Approach to Neural Network Training},
 url = {https://api.semanticscholar.org/CorpusID:226281826},
 volume = {23},
 year = {2020}
}

@inproceedings{shi2024instructiontuninglossinstructions,
 author = {Zhengxiang Shi and
Adam X. Yang and
Bin Wu and
Laurence Aitchison and
Emine Yilmaz and
Aldo Lipani},
 bibsource = {dblp computer science bibliography, https://dblp.org},
 booktitle = {Advances in Neural Information Processing Systems 38: Annual Conference
on Neural Information Processing Systems 2024, NeurIPS 2024, Vancouver,
BC, Canada, December 10 - 15, 2024},
 editor = {Amir Globersons and
Lester Mackey and
Danielle Belgrave and
Angela Fan and
Ulrich Paquet and
Jakub M. Tomczak and
Cheng Zhang},
 title = {Instruction Tuning With Loss Over Instructions},
 url = {http://papers.nips.cc/paper\_files/paper/2024/hash/7ffb43adf37b3eeaba559098bc084cc6-Abstract-Conference.html},
 year = {2024}
}

@article{RezaeiDastjerdehei2020AddressingII,
 author = {Mohammad Reza Rezaei-Dastjerdehei and Amir Mohammad Mijani and Emad Fatemizadeh},
 journal = {2020 27th National and 5th International Iranian Conference on Biomedical Engineering (ICBME)},
 pages = {333-338},
 title = {Addressing Imbalance in Multi-Label Classification Using Weighted Cross Entropy Loss Function},
 url = {https://api.semanticscholar.org/CorpusID:231684807},
 year = {2020}
}

@inproceedings{chen2024alpagasustrainingbetteralpaca,
 author = {Lichang Chen and
Shiyang Li and
Jun Yan and
Hai Wang and
Kalpa Gunaratna and
Vikas Yadav and
Zheng Tang and
Vijay Srinivasan and
Tianyi Zhou and
Heng Huang and
Hongxia Jin},
 bibsource = {dblp computer science bibliography, https://dblp.org},
 booktitle = {The Twelfth International Conference on Learning Representations,
{ICLR} 2024, Vienna, Austria, May 7-11, 2024},
 publisher = {OpenReview.net},
 title = {AlpaGasus: Training a Better Alpaca with Fewer Data},
 url = {https://openreview.net/forum?id=FdVXgSJhvz},
 year = {2024}
}

@misc{allamo,
 author = {Ociepa, Krzysztof},
 howpublished = {\url{https://github.com/chrisociepa/allamo}},
 publisher = {GitHub},
 title = {ALLaMo: A Simple, Hackable, and Fast Framework for Training Medium-Sized LLMs},
 year = {2023}
}

@inproceedings{Ansel_PyTorch_2_Faster_2024,
 author = {Ansel, Jason and Yang, Edward and He, Horace and Gimelshein, Natalia and Jain, Animesh and Voznesensky, Michael and Bao, Bin and Bell, Peter and Berard, David and Burovski, Evgeni and Chauhan, Geeta and Chourdia, Anjali and Constable, Will and Desmaison, Alban and DeVito, Zachary and Ellison, Elias and Feng, Will and Gong, Jiong and Gschwind, Michael and Hirsh, Brian and Huang, Sherlock and Kalambarkar, Kshiteej and Kirsch, Laurent and Lazos, Michael and Lezcano, Mario and Liang, Yanbo and Liang, Jason and Lu, Yinghai and Luk, CK and Maher, Bert and Pan, Yunjie and Puhrsch, Christian and Reso, Matthias and Saroufim, Mark and Siraichi, Marcos Yukio and Suk, Helen and Suo, Michael and Tillet, Phil and Wang, Eikan and Wang, Xiaodong and Wen, William and Zhang, Shunting and Zhao, Xu and Zhou, Keren and Zou, Richard and Mathews, Ajit and Chanan, Gregory and Wu, Peng and Chintala, Soumith},
 booktitle = {29th ACM International Conference on Architectural Support for Programming Languages and Operating Systems, Volume 2 (ASPLOS '24)},
 doi = {10.1145/3620665.3640366},
 publisher = {ACM},
 title = {{PyTorch 2: Faster Machine Learning Through Dynamic Python Bytecode Transformation and Graph Compilation}},
 url = {https://pytorch.org/assets/pytorch2-2.pdf},
 year = {2024}
}

@article{zhang2024tinyllamaopensourcesmalllanguage,
 author = {Peiyuan Zhang and Guangtao Zeng and Tianduo Wang and Wei Lu},
 journal = {ArXiv preprint},
 title = {TinyLlama: An Open-Source Small Language Model},
 url = {https://arxiv.org/abs/2401.02385},
 volume = {abs/2401.02385},
 year = {2024}
}

@inproceedings{dao2023flashattention2fasterattentionbetter,
 author = {Tri Dao},
 bibsource = {dblp computer science bibliography, https://dblp.org},
 booktitle = {The Twelfth International Conference on Learning Representations,
{ICLR} 2024, Vienna, Austria, May 7-11, 2024},
 publisher = {OpenReview.net},
 title = {FlashAttention-2: Faster Attention with Better Parallelism and Work
Partitioning},
 url = {https://openreview.net/forum?id=mZn2Xyh9Ec},
 year = {2024}
}

@misc{open-pl-llm-leaderboard,
 author = {Wr\'{o}bel, Krzysztof and {SpeakLeash Team} and {Cyfronet Team}},
 howpublished = {\url{https://huggingface.co/spaces/speakleash/open_pl_llm_leaderboard}},
 publisher = {Hugging Face},
 title = {Open PL LLM Leaderboard},
 year = {2024}
}

@inproceedings{zheng2023judging,
 author = {Lianmin Zheng and
Wei{-}Lin Chiang and
Ying Sheng and
Siyuan Zhuang and
Zhanghao Wu and
Yonghao Zhuang and
Zi Lin and
Zhuohan Li and
Dacheng Li and
Eric P. Xing and
Hao Zhang and
Joseph E. Gonzalez and
Ion Stoica},
 bibsource = {dblp computer science bibliography, https://dblp.org},
 booktitle = {Advances in Neural Information Processing Systems 36: Annual Conference
on Neural Information Processing Systems 2023, NeurIPS 2023, New Orleans,
LA, USA, December 10 - 16, 2023},
 editor = {Alice Oh and
Tristan Naumann and
Amir Globerson and
Kate Saenko and
Moritz Hardt and
Sergey Levine},
 title = {Judging LLM-as-a-Judge with MT-Bench and Chatbot Arena},
 url = {http://papers.nips.cc/paper\_files/paper/2023/hash/91f18a1287b398d378ef22505bf41832-Abstract-Datasets\_and\_Benchmarks.html},
 year = {2023}
}

@misc{rkinas2024mt-bench-pl,
 author = {Kinas, Remigiusz and Maria, Filipkowska and {SpeakLeash Team} and {Cyfronet Team}},
 howpublished = {\url{https://huggingface.co/spaces/speakleash/mt-bench-pl}},
 publisher = {Speakleash},
 title = {MT-Bench PL},
 year = {2024}
}

@article{frantar-gptq,
 author = {Elias Frantar and Saleh Ashkboos and Torsten Hoefler and Dan Alistarh},
 journal = {ArXiv preprint},
 title = {{GPTQ}: Accurate Post-training Compression for Generative Pretrained Transformers},
 url = {https://arxiv.org/abs/2210.17323},
 volume = {abs/2210.17323},
 year = {2022}
}

@misc{eval-harness,
 author = {Gao, Leo and Tow, Jonathan and Abbasi, Baber and Biderman, Stella and Black, Sid and DiPofi, Anthony and Foster, Charles and Golding, Laurence and Hsu, Jeffrey and Le Noac'h, Alain and Li, Haonan and McDonell, Kyle and Muennighoff, Niklas and Ociepa, Chris and Phang, Jason and Reynolds, Laria and Schoelkopf, Hailey and Skowron, Aviya and Sutawika, Lintang and Tang, Eric and Thite, Anish and Wang, Ben and Wang, Kevin and Zou, Andy},
 doi = {10.5281/zenodo.12608602},
 publisher = {Zenodo},
 title = {A framework for few-shot language model evaluation},
 url = {https://zenodo.org/records/12608602},
 version = {v0.4.3},
 year = {2024}
}

@inproceedings{kocon-etal-2019-multi,
 address = {Hong Kong, China},
 author = {Koco{\'n}, Jan  and
Mi{\l}kowski, Piotr  and
Za{\'s}ko-Zieli{\'n}ska, Monika},
 booktitle = {Proceedings of the 23rd Conference on Computational Natural Language Learning (CoNLL)},
 doi = {10.18653/v1/K19-1092},
 editor = {Bansal, Mohit  and
Villavicencio, Aline},
 pages = {980--991},
 publisher = {Association for Computational Linguistics},
 title = {Multi-Level Sentiment Analysis of {P}ol{E}mo 2.0: Extended Corpus of Multi-Domain Consumer Reviews},
 url = {https://aclanthology.org/K19-1092},
 year = {2019}
}

@inproceedings{marcinczuk2013open,
 author = {Marcinczuk, Micha{\l} and Ptak, Marcin and Radziszewski, Adam and Piasecki, Maciej},
 booktitle = {Proceedings of the 6th Language \& Technology Conference: Human Language Technologies as a Challenge for Computer Science and Linguistics, Wydawnictwo Poznanskie, Fundacja Uniwersytetu im. Adama Mickiewicza},
 title = {Open dataset for development of Polish Question Answering systems},
 year = {2013}
}

@inproceedings{ogro:kop:14:lrec,
 address = {Reykjavik, Iceland},
 author = {Ogrodniczuk, Maciej  and
Kope{\'c}, Mateusz},
 booktitle = {Proceedings of the Ninth International Conference on Language Resources and Evaluation ({LREC}'14)},
 editor = {Calzolari, Nicoletta  and
Choukri, Khalid  and
Declerck, Thierry  and
Loftsson, Hrafn  and
Maegaard, Bente  and
Mariani, Joseph  and
Moreno, Asuncion  and
Odijk, Jan  and
Piperidis, Stelios},
 pages = {3712--3715},
 publisher = {European Language Resources Association (ELRA)},
 title = {The {P}olish Summaries Corpus},
 url = {http://www.lrec-conf.org/proceedings/lrec2014/pdf/1211_Paper.pdf},
 year = {2014}
}

@inproceedings{dadas-etal-2020-evaluation,
 address = {Marseille, France},
 author = {Dadas, Slawomir  and
Pere{\l}kiewicz, Micha{\l}  and
Po{\'s}wiata, Rafa{\l}},
 booktitle = {Proceedings of the Twelfth Language Resources and Evaluation Conference},
 editor = {Calzolari, Nicoletta  and
B{\'e}chet, Fr{\'e}d{\'e}ric  and
Blache, Philippe  and
Choukri, Khalid  and
Cieri, Christopher  and
Declerck, Thierry  and
Goggi, Sara  and
Isahara, Hitoshi  and
Maegaard, Bente  and
Mariani, Joseph  and
Mazo, H{\'e}l{\`e}ne  and
Moreno, Asuncion  and
Odijk, Jan  and
Piperidis, Stelios},
 isbn = {979-10-95546-34-4},
 language = {English},
 pages = {1674--1680},
 publisher = {European Language Resources Association},
 title = {Evaluation of Sentence Representations in {P}olish},
 url = {https://aclanthology.org/2020.lrec-1.207},
 year = {2020}
}

@inproceedings{9945218,
 author = {Dadas, S{\l}awomir},
 booktitle = {2022 IEEE International Conference on Systems, Man, and Cybernetics (SMC)},
 doi = {10.1109/SMC53654.2022.9945218},
 number = {},
 pages = {371-378},
 title = {Training Effective Neural Sentence Encoders from Automatically Mined Paraphrases},
 volume = {},
 year = {2022}
}

@article{ptaszynski2023expert,
 author = {Ptaszynski, Michal and Pieciukiewicz, Agata and Dybala, Pawel and Skrzek, Pawel and Soliwoda, Kamil and Fortuna, Marcin and Leliwa, Gniewosz and Wroczynski, Michal},
 journal = {Data},
 number = {1},
 pages = {1},
 publisher = {MDPI},
 title = {Expert-Annotated Dataset to Study Cyberbullying in Polish Language},
 volume = {9},
 year = {2023}
}

@inproceedings{rybak-etal-2024-polqa-polish,
 address = {Torino, Italia},
 author = {Rybak, Piotr  and
Przyby{\l}a, Piotr  and
Ogrodniczuk, Maciej},
 booktitle = {Proceedings of the 2024 Joint International Conference on Computational Linguistics, Language Resources and Evaluation (LREC-COLING 2024)},
 editor = {Calzolari, Nicoletta  and
Kan, Min-Yen  and
Hoste, Veronique  and
Lenci, Alessandro  and
Sakti, Sakriani  and
Xue, Nianwen},
 pages = {12846--12855},
 publisher = {ELRA and ICCL},
 title = {{P}ol{QA}: {P}olish Question Answering Dataset},
 url = {https://aclanthology.org/2024.lrec-main.1125},
 year = {2024}
}

@inproceedings{tuora2023poquad,
 author = {Tuora, Ryszard and Zwierzchowska, Aleksandra and Zawadzka-Paluektau, Natalia and Klamra, Cezary and Kobyli{\'n}ski, {\L}ukasz},
 booktitle = {Proceedings of the 12th Knowledge Capture Conference 2023},
 pages = {105--113},
 title = {PoQuAD-The Polish Question Answering Dataset-Description and Analysis},
 year = {2023}
}

@inproceedings{bandarkar-etal-2024-belebele,
 address = {Bangkok, Thailand and virtual meeting},
 author = {Bandarkar, Lucas  and
Liang, Davis  and
Muller, Benjamin  and
Artetxe, Mikel  and
Shukla, Satya Narayan  and
Husa, Donald  and
Goyal, Naman  and
Krishnan, Abhinandan  and
Zettlemoyer, Luke  and
Khabsa, Madian},
 booktitle = {Proceedings of the 62nd Annual Meeting of the Association for Computational Linguistics (Volume 1: Long Papers)},
 pages = {749--775},
 publisher = {Association for Computational Linguistics},
 title = {The Belebele Benchmark: a Parallel Reading Comprehension Dataset in 122 Language Variants},
 url = {https://aclanthology.org/2024.acl-long.44},
 year = {2024}
}

@inproceedings{rybak-etal-2020-klej,
 address = {Online},
 author = {Rybak, Piotr  and
Mroczkowski, Robert  and
Tracz, Janusz  and
Gawlik, Ireneusz},
 booktitle = {Proceedings of the 58th Annual Meeting of the Association for Computational Linguistics},
 doi = {10.18653/v1/2020.acl-main.111},
 editor = {Jurafsky, Dan  and
Chai, Joyce  and
Schluter, Natalie  and
Tetreault, Joel},
 pages = {1191--1201},
 publisher = {Association for Computational Linguistics},
 title = {{KLEJ}: Comprehensive Benchmark for {P}olish Language Understanding},
 url = {https://aclanthology.org/2020.acl-main.111},
 year = {2020}
}

@misc{open-llm-leaderboard-v1,
 author = {Edward Beeching and Clémentine Fourrier and Nathan Habib and Sheon Han and Nathan Lambert and Nazneen Rajani and Omar Sanseviero and Lewis Tunstall and Thomas Wolf},
 howpublished = {\url{https://huggingface.co/spaces/open-llm-leaderboard-old/open_llm_leaderboard}},
 publisher = {Hugging Face},
 title = {Open LLM Leaderboard (2023-2024)},
 year = {2023}
}

@misc{badri2023hqq,
 author = {Hicham Badri and Appu Shaji},
 title = {Half-Quadratic Quantization of Large Machine Learning Models},
 url = {https://mobiusml.github.io/hqq_blog/},
 year = {2023}
}

@inproceedings{lin2023awq,
 author = {Lin, Ji and Tang, Jiaming and Tang, Haotian and Yang, Shang and Chen, Wei-Ming and Wang, Wei-Chen and Xiao, Guangxuan and Dang, Xingyu and Gan, Chuang and Han, Song},
 booktitle = {MLSys},
 title = {AWQ: Activation-aware Weight Quantization for LLM Compression and Acceleration},
 year = {2024}
}

@misc{mlx2023,
 author = {Awni Hannun and Jagrit Digani and Angelos Katharopoulos and Ronan Collobert},
 title = {{MLX}: Efficient and flexible machine learning on Apple silicon},
 url = {https://github.com/ml-explore},
 version = {0.0},
 year = {2023}
}

@misc{qra,
 author = { {National Information Processing Institute and Gdańsk University of Technology} },
 publisher = { Hugging Face },
 title = { {Qra models} },
 url = { https://huggingface.co/OPI-PG },
 year = {2024}
}

@misc{trurl,
 author = { {Voicelab} },
 publisher = { Hugging Face },
 title = { TRURL 2 models) },
 url = { https://huggingface.co/Voicelab },
 year = {2023}
}

\clearpage

\appendix

\section{Examples of Tasks}

\subsection{polemo2}
\textbf{Task:} Sentiment analysis of online consumer reviews across four domains (medicine, hotels, products, university) with four-class labeling (positive, negative, neutral, ambiguous).

\textbf{Example:}

\textit{Opinia: "Leczyłam się u niej parę lat i nic mi nie pomogła, a jak zmieniła m lekarza po krótkim czasie zobaczyła m już poprawę a w tej chwili jestem już bez leków 6 lat i jest wszystko dobrze . Dr Ciborska leczyła mnie na depresję a potem przez dr Kopystecką miała m rozpoznany zespół maniakalno depresyjny i odtąd zmianę leków i przede wszystkim wysłuchała mnie z zaangażowaniem a nie jak dr Ciborska aby mnie zbyć"}

\textit{Określ sentyment podanej opinii. Możliwe odpowiedzi:}

\textit{A - Neutralny, B - Negatywny, C - Pozytywny, D - Niejednoznaczny}

\textit{Prawidłowa odpowiedź: B}

\subsection{klej-ner}
\textbf{Task:} Named entity recognition in sentences containing single-type entities, classifying into six categories (no entity, place, person, organization, time, geographical name).

\textbf{Example:}

\textit{Zdanie: "Ulewne deszcze nawiedziły także Słupsk ."}

\textit{Pytanie: Jakiego rodzaju jest nazwana jednostka, jeżeli występuje w podanym zdaniu?}

\textit{Możliwe odpowiedzi: A - Brak nazwanej jednostki, B - Nazwa miejsca, C - Nazwa osoby, D - Nazwa organizacji, E - Czas, F - Nazwa geograficzna}

\textit{Prawidłowa odpowiedź: B}

\subsection{8tags}
\textbf{Task:} Topic classification of social media headlines into eight categories (film, history, food, medicine, motorization, work, sport, technology).

\textbf{Example:}

\textit{Tytuł: "Czy bateria zrobiona z 1000 cytryn jest w stanie uruchomić silnik?."}

\textit{Pytanie: jaka kategoria najlepiej pasuje do podanego tytułu?}

\textit{Możliwe odpowiedzi:}

\textit{A - film, B - historia, C - jedzenie, D - medycyna, E - motoryzacja, F - praca, G - sport, H - technologie}

\textit{Prawidłowa odpowiedź: E}

\subsection{belebele}
\textbf{Task:} Machine reading comprehension for question answering.

\textbf{Example:}

\textit{Fragment: "Atom może być uważany za jeden z fundamentalnych elementów budujących całą materię. To bardzo złożona jednostka, która składa się, według uproszczonego modelu Bohra, z centralnego jądra, wokół którego znajdują się elektrony, co nieco przypomina planety krążące wokół Słońca – patrz rysunek 1.1. W skład jądra wchodzą dwa typy cząsteczek: neutrony i protony. Pod względem ładunku elektrycznego protony są dodatnie, elektrony są ujemne, a neutrony nie mają żadnego ładunku."}

\textit{Pytanie: "Jaki ładunek mają cząstki krążące wokół jądra?"}

\textit{Możliwe odpowiedzi: A - Ładunek dodatni, B - Bez ładunku, C - Ładunek ujemny, D - Ładunek dodatni i ujemny}

\textit{Prawidłowa odpowiedź: C}

\subsection{dyk}
\textbf{Task:} Question answering based on human-annotated pairs from Wikipedia's "Did You Know" section.

\textbf{Example:}

\textit{Pytanie: "za co Iwanowi Tyszkiwiczowi ucięto dłoń?"}

\textit{Sugerowana odpowiedź: "Tyszkiewicz był torturowany – wyrwano mu język za "bluźnierstwo przeciw Bogu", a za rzucenie krucyfiksu na ziemię ucięto mu dłoń i nogę."}

\textit{Czy sugerowana odpowiedź na zadane pytanie jest poprawna? Możliwe opcje:
A - brakuje sugerowanej odpowiedzi, B - nie, sugerowana odpowiedź nie jest poprawna, C - tak, sugerowana odpowiedź jest poprawna, D - brakuje pytania}

\textit{Prawidłowa opcja: C}

\subsection{ppc}
\textbf{Task:} Text similarity assessment using manually labeled sentence pairs (exact paraphrases, close paraphrases, non-paraphrases).

\textbf{Example:}

\textit{Zdanie A: "Piasek nad Chinami."}

\textit{Zdanie B: "Burza piaskowa w Chinach."}

\textit{Pytanie: jaka jest zależność między zdaniami A i B?}

\textit{Możliwe odpowiedzi: A - wszystkie odpowiedzi poprawne, B - znaczą dokładnie to samo, C - mają podobne znaczenie, D - mają różne znaczenie}

\textit{Prawidłowa odpowiedź: C}

\subsection{psc}
\textbf{Task:} Summarization of news articles.

\textbf{Example:}

\textit{Fragment 1: "Zwykle zaczyna się od siąkających nosami kilku osób. Jednak choroba postępuje lawinowo. Wirus grypy przenosi się drogą kropelkową - podczas rozmowy, kaszlu i kichania. Jedna zagrypiona osoba, która pojawi się w towarzystwie, może zakazić wielu ludzi.Lekarz dyżurny kraju Michał Sobolewski uspokaja: w Polsce jeszcze nie ma epidemii grypy. Wybuchnie, kiedy będzie dużo źródeł zakażenia."}

\textit{Fragment 2: "W niedzielę przychodnie w Warszwie zalała fala pacjentów z objawami grypy. Lekarz dyżurny kraju uspokaja, że w Polsce nie ma jeszcze epidemii grypy. Podkreśla też, że Polacy lekceważą profilaktyczne szczepienia przeciwko tej chorobie, a tylko one zapobiegają rozprzestrzenianiu się schorzeń zakaźnych. W Europie epidemia grypy dociera do kolejnych państw. Odnotowano już przypadki śmiertelne."}

\textit{Pytanie: jaka jest zależność między fragmentami 1 i 2?}

\textit{Możliwe odpowiedzi: A - wszystkie odpowiedzi poprawne, B - dotyczą tego samego artykułu, C - dotyczą różnych artykułów, D - brak poprawnej odpowiedzi}

\textit{Prawidłowa odpowiedź: B}

\subsection{cbd}
\textbf{Task:} Text classification for cyberbullying and hate-speech detection.

\textbf{Example:}

\textit{Wypowiedź: "Ty wiesz lepiej. Ja wiem, że nawet wiceprezydentem nie będziesz na 100\%"}

\textit{Pytanie: Jaka kategoria najlepiej pasuje do podanej wypowiedzi?}

\textit{Możliwe odpowiedzi: A - nieszkodliwa, B - szyderstwo, C - obelga, D - insynuacja, E - groźba, F - molestowanie}

\textit{Prawidłowa odpowiedź: B}

\subsection{polqa}
\textbf{Task:} Open-domain question answering from the "Jeden z dziesięciu" TV show, with and without context (abstractive QA/RAG).

\textbf{Example:}

\textit{Kontekst: Przymiotnik. Przymiotniki, podobnie jak w języku polskim, odmieniały się przez liczby, rodzaje i przypadki. Wyraz określający następował zawsze po wyrazie określanym, tak jak w innych językach semickich, np. "ezzūtu šārū" „porywiste wiatry", dosłownie „wiatry porywiste"}

\textit{Pytanie: Czy przymiotniki odmienia się przez przypadki?}

\textit{Czy kontekst jest relewantny dla pytania?}

\textit{Odpowiedz krótko "Tak" lub "Nie". Prawidłowa odpowiedź: Tak}

\textit{Kontekst: Alibi (łac. gdzie indziej) – dowód w postępowaniu karnym na okoliczność, że podejrzany albo oskarżony znajdował się w miejscu innym niż miejsce popełnienia zarzucanego mu przestępstwa.}

\textit{Pytanie: Jak z łaciny nazywa się dowód sądowy polegający na wykazaniu, że osoba oskarżona nie przebywała na miejscu przestępstwa w chwili gdy je popełniono?}

\textit{Prawidłowa odpowiedź: alibi}

\subsection{poquad}
\textbf{Task:} Context-based extractive question answering (QA/RAG).
\textbf{Example:}

\textit{Tytuł: Miszna}

\textit{Kontekst: Pisma rabiniczne – w tym Miszna – stanowią kompilację poglądów różnych rabinów na określony temat. Zgodnie z wierzeniami judaizmu Mojżesz otrzymał od Boga całą Torę, ale w dwóch częściach: jedną część w formie pisanej, a drugą część w formie ustnej. Miszna – jako Tora ustna – była traktowana nie tylko jako uzupełnienie Tory spisanej, ale również jako jej interpretacja i wyjaśnienie w konkretnych sytuacjach życiowych. Tym samym Miszna stanowiąca kodeks Prawa religijnego zaczęła równocześnie służyć za jego ustnie przekazywany podręcznik.}

\textit{Pytanie: Czym są pisma rabiniczne?}

\textit{Prawidłowa odpowiedź (krótki cytat z Kontekstu): kompilację poglądów różnych rabinów na określony temat}

\section{Evaluation Reproducibility}

To reproduce our results, you need to clone the repository:

\begin{lstlisting}[language=bash, basicstyle=\small]
git clone https://github.com/speakleash/lm-evaluation-harness.git -b polish3
cd lm-evaluation-harness
pip install -e .
\end{lstlisting}

and run benchmark for 0-shot and 5-shot:

\begin{lstlisting}[language=bash, basicstyle=\small]
lm_eval --model hf --model_args pretrained=speakleash/Bielik-7B-Instruct-v0.1 --tasks polish_generate --num_fewshot 0 --output_path results/ --log_samples

lm_eval --model hf --model_args pretrained=speakleash/Bielik-7B-Instruct-v0.1 --tasks polish_mc --num_fewshot 0 --output_path results/ --log_samples

lm_eval --model hf --model_args pretrained=speakleash/Bielik-7B-Instruct-v0.1 --tasks polish_generate_few --num_fewshot 5 --output_path results/ --log_samples

lm_eval --model hf --model_args pretrained=speakleash/Bielik-7B-Instruct-v0.1 --tasks polish_mc --num_fewshot 5 --output_path results/ --log_samples
\end{lstlisting}

\end{document}